\documentclass[journal]{IEEEtran}

\usepackage{cite}
\usepackage{float}
\usepackage{amsmath,amssymb}

\ifCLASSINFOpdf
  \usepackage[pdftex]{graphicx}
\else
\fi

\begin{document}
\bstctlcite{ralReferenceStyle}

\title{TEOD-Planner: Tracking-Error-Aware Whole-Body Planning for USVs in Narrow Waterways}

\author{Tianren~Liu, Xufei~Cui, Jiankun~Yang, Shaobo~Zhou, Hua~Zhu,
        Boyu~Zhou, Fei~Gao, Zhiyun~Lin, and Haibo~Lu}

\maketitle

\begin{abstract}
Safe navigation through narrow waterways requires trajectory planners to dynamically adjust obstacle clearance in response to tracking errors while ensuring rapid convergence. However existing local planners commonly approximate the robot geometry through isotropic inflation and further increase the inflation margin to account for tracking errors, leading to conservative motion in tight spaces. To address the challenge, this work proposes a trajectory planning framework based on Tracking-Error-oriented Occupancy Deformation (TEOD). First, a spatio-temporal convolutional network is employed to predict future position and yaw tracking errors. Then, TEOD maps these predicted errors to deformations of the vessel's whole-body occupancy, accounting for its shape and heading. This deformation enables the planner to adapt safety margins to predicted tracking errors while considering the vessel's heading-dependent geometry. Finally, the planner incorporates the resulting occupancy representation into trajectory optimization and uses analytical gradients to support efficient joint planning of position and heading. Simulation results demonstrate that the proposed framework avoids hull--obstacle contact across all 90 scenarios with low, medium, and high obstacle densities. Real-world experiments further demonstrate collision-free navigation over five consecutive traversals of a complex obstacle course, with an average onboard optimization time of approximately $5\,\mathrm{ms}$.
\end{abstract}


\IEEEpeerreviewmaketitle

\section{Introduction}

Unmanned surface vessels (USVs) are increasingly deployed for inland waterway transportation\cite{barrera2021trends}, surveys\cite{zhao2024optimal}, and rescue~\cite{vagale2021asvReview}.
In practical long-range navigation, however, a USV must traverse narrow, cluttered waterways and contend with persistent water-current disturbances. This setting makes local planning challenging: the planner must account for the yaw-dependent body footprint in tight spaces~\cite{zheng_scan-planner_2026} and leave sufficient clearance to accommodate the resulting tracking errors~\cite{lekkas2016ilos}.

\begin{figure}[t]
\centering
\includegraphics[width=\columnwidth]{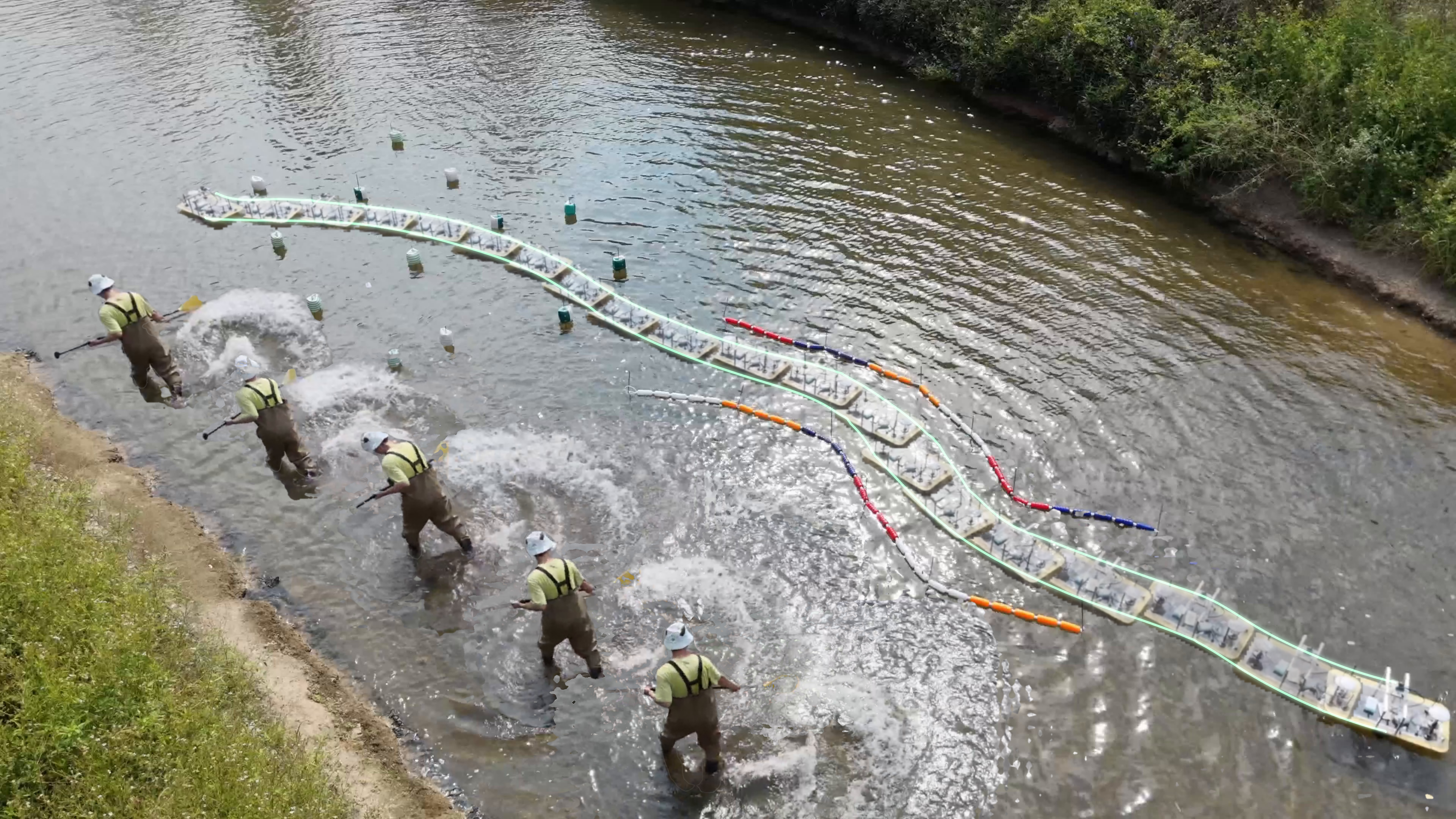}
\caption{Real-world USV navigation through a narrow obstacle course.}
\label{fig:traj_comparison}
\end{figure}

Numerous studies have investigated planning and control methods to address this challenge.
Residual learning-based model predictive control (MPC) methods~\cite{gamboa-gonzalez_online_2025} are commonly used to improve trajectory tracking under external disturbances. However, uncertainties in residual predictions may still compromise tracking reliability. Stochastic MPC addresses state uncertainty through uncertainty propagation and chance constraints for path tracking and obstacle avoidance~\cite{park_robust_2023}. Although computationally cheap, neglecting the yaw-dependent footprint makes these representations overly conservative in tight spaces~\cite{zheng_scan-planner_2026}. Whole-body planners have been developed to improve collision fidelity. For example, RC-ESDF~\cite{geng_robo-centric_2023} uses a robot-centric Euclidean signed distance field to evaluate whole-body collisions for robots of arbitrary shapes. 3D RoA-Planner searches collision-free yaw ranges to generate feasible paths~\cite{son_3d_2026}. Nevertheless, explicit yaw-state reasoning increases planning complexity. To address this issue, SCAN-Planner uses trajectory-induced heading for whole-body collision avoidance without independent yaw optimization~\cite{zheng_scan-planner_2026}. However, SCAN-Planner does not explicitly account for tracking errors during trajectory planning. Therefore, existing methods rarely account for both tracking errors and whole-body occupancy during trajectory planning. Directly combining these approaches introduces substantial computational overhead, making real-time optimization challenging. 

To address these challenges, we propose TEOD-Planner for whole-body trajectory planning.
First, a spatio-temporal convolutional network (STCN) predicts future sequences of position and yaw tracking errors from execution history and a candidate trajectory.
Next, TEOD maps these predicted errors into whole-body occupancy deformation based on the vessel’s shape and heading.
Finally, the deformed occupancy is incorporated into trajectory
optimization with analytical clearance gradients for efficient
joint position–heading planning. 
Our method computes safety margins from the deformed occupancy to enable efficient trajectory optimization, as shown in Fig.~\ref{fig:traj_comparison}. Across 90 simulated navigation scenes and real-world trials, TEOD recorded zero hull--obstacle contacts with an average onboard optimization time of approximately $5\,\mathrm{ms}$. The contributions of this paper are as follows:
\begin{enumerate}
\item To overcome the limitations of conservative fixed safety margins, we design STCN that predicts future position and yaw tracking-error sequences from history trajectories, enabling adaptive clearance estimation.

\item To simplify the coupled constraints of tracking errors and whole-body geometry, we propose TEOD, which unifies tracking error and the vessel footprint into an exact deformed occupancy region.

\item To ensure real-time optimization, we derive analytical clearance gradients incorporating position and heading uncertainty for joint trajectory planning. Extensive simulations and real-world experiments confirm zero contacts at an average onboard solve time of approximately $5\,\mathrm{ms}$.
\end{enumerate}
\begin{figure*}[!t]
\centering
\includegraphics[width=\textwidth]{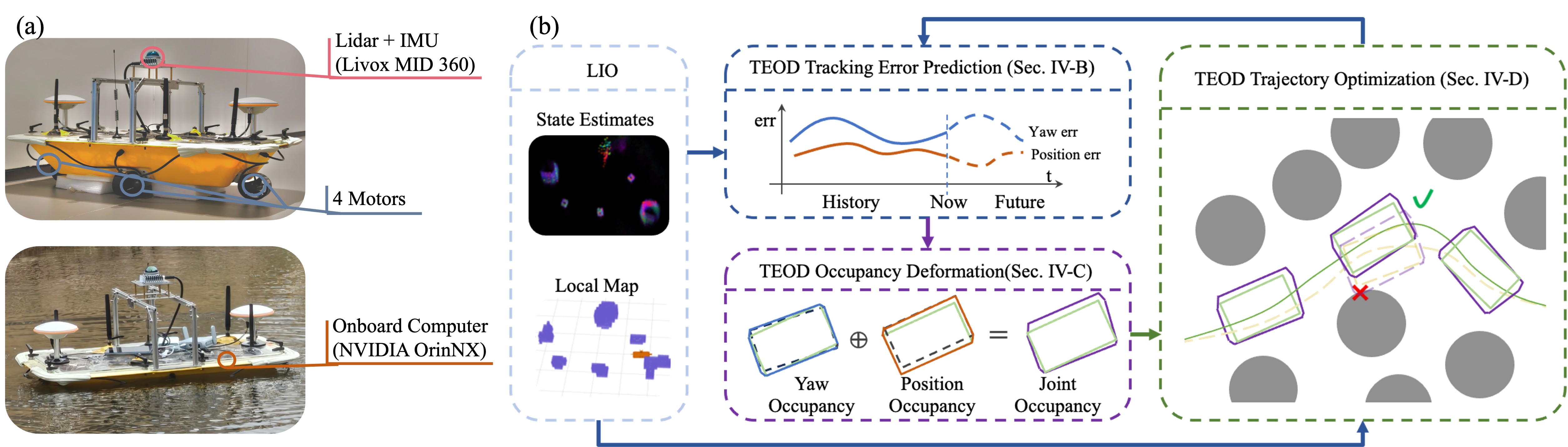}
\caption{System framework. (a) USV platform and hardware installation.
(b) Prediction, occupancy construction, and trajectory optimization.
First, predicted position and yaw errors jointly deform vessel occupancy.
Then, the optimizer uses this occupancy and the local map to update the trajectory.}
\label{fig:system_architecture}
\end{figure*}

\section{Related Work}

Existing research on disturbance-aware marine planning is reviewed in this section. A survey of whole-body trajectory planning methods is also presented.

\subsection{Tracking errors Marine Planning}

Numerous studies improve marine trajectory tracking through disturbance estimation, model learning, and disturbance preview. For example, van der Saag et al.~\cite{van_der_saag_active_2025} adopt active disturbance rejection control, using an extended state observer to estimate disturbances for feedforward compensation. In addition, Gamboa-Gonzalez et al.~\cite{gamboa-gonzalez_online_2025} integrate online residual learning into MPC to account for unmodeled dynamics and environmental disturbances. Duan et al.~\cite{duan_multimode_2025} use an LSTM-assisted MPC scheme for USV tracking across operating conditions. To anticipate wave disturbances, Walker and Giorgio-Serchi~\cite{walker_disturbance_2023} incorporate predicted wave loads into nonlinear MPC for underwater vehicle tracking. At the planning level, Akbari et al.~\cite{akbari2025currentGPMP} incorporate predicted currents into Gaussian process motion planning. Robust and stochastic predictive controllers also account for uncertainty during tracking. For example, Tagliabue et al.~\cite{tagliabue2023robust} train a neural network policy to approximate robust tube MPC for trajectory tracking of insect-scale aerial robots. For autonomous ships, Park et al.~\cite{park_robust_2023} adopt stochastic MPC for path tracking and obstacle avoidance under uncertainty. Wang et al.~\cite{wang_parametric_2025} incorporate disturbance estimates and collision avoidance constraints into nonlinear MPC for USV tracking. In addition, Wang et al.~\cite{wang2024mpccbf} combine MPC with control barrier functions to improve disturbance rejection while keeping vessel motion within a predefined safe region. However, conservative uncertainty bounds or safety margins can restrict the available planning space in narrow waterways.

\subsection{Whole-body Trajectory Planning}

Local planners generate trajectories through corridor-constrained optimization, such as GCOPTER~\cite{wang_geometrically_2022}, or gradient-based replanning, such as EGO-Planner~\cite{zhou2021egoPlanner} and EGO-v2~\cite{zhou_swarm_2022}. However, these planners typically approximate the robot's geometry using a spherical bounding volume, resulting in isotropic safety margins that neglect heading-dependent body occupancy. To achieve whole-body collision avoidance, Geng et al.~\cite{geng_robo-centric_2023} propose RC-ESDF, which uses a robot-centric distance field to jointly optimize position and orientation. Zhou and Kong~\cite{zhou_likelihood_2025} account for rotation when initializing trajectories for polygonal vehicles. Jaitly et al.~\cite{jaitly_analytic_2025} approximate polytopes by smooth sets to derive differentiable nonpenetration constraints. To reduce the complexity of position-orientation search, Son et al.~\cite{son_3d_2026} introduce 3D RoA-Planner, which identifies collision-free and stable yaw ranges for reduced-dimensional path planning. To improve planning efficiency, Zheng et al.~\cite{zheng_scan-planner_2026} develop SCAN-Planner, which couples robot heading with trajectory geometry to enable whole-body collision avoidance without independent yaw optimization. For USV navigation, Gonzalez-Garcia et al.~\cite{gonzalez-garcia_safe_2025} combine MPC with high-order control barrier functions, adapting obstacle inflation to the vessel's relative position and orientation. These methods factor in body shape and orientation, avoiding the overly conservative results of isotropic approximations. However, coupling tracking errors with whole-body geometric constraints makes optimization hard to converge in narrow waterways.

\section{System Architecture}
\label{sec:system_architecture}

\subsection{USV Platform}

Fig.~\ref{fig:system_architecture}(a) shows the USV platform and hardware installation.
Specifically, our USV is equipped with four thrusters arranged in a cross-shaped configuration~\cite{wang2018design}.
A Livox MID-360 LiDAR provides range and inertial measurements for localization and mapping.
An NVIDIA Jetson Orin NX executes the navigation modules onboard.

We model the planar motion of the USV using three-degree-of-freedom dynamics~\cite{wang2018design}:
\begin{equation}
    \mathbf M\dot{\mathbf v}_{\mathrm b}
    +\mathbf C(\mathbf v_{\mathrm b})\mathbf v_{\mathrm b}
    +\mathbf D(\mathbf v_{\mathrm b})\mathbf v_{\mathrm b}
    =\boldsymbol\tau,
    \label{eq:usv_dynamics}
\end{equation}
where $\mathbf v_{\mathrm b}=[u,v,r]^{\top}$ contains the body-frame surge velocity, sway velocity, and yaw rate. The matrices $\mathbf M$, $\mathbf C(\mathbf v_{\mathrm b})$, and $\mathbf D(\mathbf v_{\mathrm b})\in\mathbb R^{3\times3}$ represent total inertia, Coriolis--centripetal effects, and hydrodynamic damping, respectively. The vector $\boldsymbol\tau\in\mathbb R^3$ contains the commanded surge force, sway force, and yaw moment.

\subsection{Software Architecture}

Fig.~\ref{fig:system_architecture}(b) illustrates the software pipeline of TEOD-Planner. Initially, LiDAR-inertial odometry (LIO) provides vessel state estimates and a local obstacle map. Then, the prediction module collects historical position and yaw tracking errors from the executed and reference trajectories. Subsequently, STCN predicts future tracking errors along the candidate trajectory (Sec.~\ref{subsec:residual_prediction}). Next, TEOD maps these predictions into whole-body occupancy deformation according to the vessel's shape and heading (Sec.~\ref{subsec:residual_deformation}). The deformed occupancy is incorporated into trajectory optimization to generate a collision-aware reference trajectory (Sec.~\ref{subsec:deformable_trajectory_optimization}). Finally, a disturbance-rejection controller tracks the reference trajectory, and the resulting tracking errors are fed back into the prediction module for the next planning cycle.

\section{Tracking-Error-Driven Occupancy Deformation Planner}
\label{sec:method}
\subsection{Problem Formulation}
\label{subsec:problem_formulation}

The planner seeks a smooth trajectory with clearance for the predicted vessel occupancy.
Position and yaw tracking errors change the hull's location and orientation.
Execution history and candidate motion determine a joint position--yaw error set $\mathcal E(t)$ through prediction and calibration.
The hull geometry and nominal heading map $\mathcal E(t)$ into an occupancy enclosure $\mathcal K(t)$ relative to the reference position.
The candidate trajectory determines both sets.

MINCO~\cite{wang_geometrically_2022} parameterizes the planar trajectory $\mathbf p(t)$ with intermediate waypoints $\mathbf q$ and segment durations $\mathbf T$.
The trajectory tangent determines nominal yaw $\psi(t)$.
The parameterization incorporates the prescribed endpoint states and segment continuity.
The planning problem balances squared jerk and duration under motion, hull-enclosure, and clearance constraints:
\begin{equation}
\begin{aligned}
    \min_{\mathbf q,\mathbf T}\quad
    &J=\int_0^{T_\Sigma}\|\mathbf p^{(3)}(t)\|_2^2\,dt+w_TT_\Sigma,\\
    \mathrm{s.t.}\quad
    &g_\ell(\mathbf p^{(\ell)}(t))\leq0,
      \quad \ell=0,1,2,3,\\
    &\mathbf r+\mathbf R(\psi(t)+\varepsilon)\mathcal H
      \subseteq\mathcal K(t),\\
    &\qquad\forall(\mathbf r,\varepsilon)\in\mathcal E(t),\\
    &\operatorname{dist}\!\left(\mathbf p(t)+\mathcal K(t),
        \mathcal O\right)\geq\delta_{\mathrm h},\\
    &T_i>0,\quad i=1,\ldots,M.
\end{aligned}
\label{eq:rod_problem}
\end{equation}
Here, $T_\Sigma=\sum_{i=1}^{M}T_i$ is the total duration, and $w_T>0$ weights travel time.
All time-dependent constraints apply over $t\in[0,T_\Sigma]$.
The function $g_\ell$ specifies the motion constraint for derivative order $\ell$.
The vector $\mathbf r$ and angle $\varepsilon$ denote position and yaw tracking errors.
The set $\mathcal H$ is the body-frame hull, and $\mathbf R$ denotes planar rotation.
The set $\mathcal O$ contains raw-map obstacles.
The operator $\operatorname{dist}$ gives the minimum Euclidean distance between two sets, and $\delta_{\mathrm h}>0$ is the required hull clearance.

The main difficulty in solving this problem lies in the nonlinear coupling between the trajectory variables and the clearance constraints. This coupling complicates gradient evaluation and hinders the rapid convergence of trajectory optimization. To address this difficulty, we use a convex hull to capture the combined effects of position and yaw tracking errors. This representation enables the analytical derivation of clearance gradients by explicitly describing how the hull extent along obstacle directions varies with the trajectory variables. Sec.~\ref{subsec:residual_prediction} uses the STCN to predict position and yaw tracking errors and calibrates their bounds, Sec.~\ref{subsec:residual_deformation} uses these bounds to construct a convex hull enclosing the possible vessel footprints, and Sec.~\ref{subsec:deformable_trajectory_optimization} derives analytical clearance gradients from this hull for trajectory optimization.

\subsection{Tracking Error Prediction}
\label{subsec:residual_prediction}

Considering position tracking errors alone cannot accurately estimate hull--obstacle clearance. Accounting for both position and yaw tracking errors enables a more accurate estimation of the vessel footprint and its clearance from obstacles. Therefore, we use an STCN to jointly predict both tracking errors from the vessel's recent motion and tracking-error history, conditioned on the candidate trajectory:
\begin{equation}
    \{\hat{\boldsymbol\xi}_h\}_{h=1}^{H}
    =f_{\boldsymbol\theta}\!\left(
    \mathcal X_{\mathrm{act}}^{-},
    \mathcal X_{\mathrm{plan}}^{+},\mathcal R^{-}\right),
    \label{eq:residual_prediction}
\end{equation}
Here, $\mathcal X_{\mathrm{act}}^{-}$ and $\mathcal R^{-}$ denote motion and error histories over the past $3\,\mathrm{s}$.
The input $\mathcal X_{\mathrm{plan}}^{+}$ describes the candidate motion over the next $3\,\mathrm{s}$.
The network $f_{\boldsymbol\theta}$ outputs $\hat{\boldsymbol\xi}_h=[\hat{\mathbf r}_h^T,\hat\varepsilon_h]^T$ for a prediction horizon of $h$ seconds, where $h=1,\ldots,H$ and $H=3$.
The vector $\hat{\mathbf r}_h$ predicts actual-minus-reference position error in world axes.
The scalar $\hat\varepsilon_h$ predicts signed yaw error relative to the trajectory tangent in $[-\pi,\pi)$.
A single forward pass produces all three predictions.

Prediction uncertainty requires error ranges around these estimates.
A polygon and a circular arc represent the position and yaw ranges:
\begin{equation}
\begin{aligned}
    \mathcal U_h
    &=\{\mathbf d:\boldsymbol\nu_m^T\mathbf d\leq b_{h,m},
        \ m=1,\ldots,D\},\\
    \mathcal E_h
    &=(\hat{\mathbf r}_h+\mathcal U_h)
      \times\operatorname{arc}(\hat\varepsilon_h,b_{h,\psi}).
    \label{eq:calibrated_residual_polygon}
\end{aligned}
\end{equation}
Here, $\boldsymbol\nu_m$ denotes one of $D=8$ symmetric world-frame unit directions.
The widths $b_{h,m}$ and $b_{h,\psi}$ bound directional position deviations and circular yaw deviations, respectively.
A yaw half-width of at least $\pi$ gives the full circle.
The symbols $\mathcal E_{h,\mathrm p}$ and $\mathcal E_{h,\psi}$ denote the two factors of $\mathcal E_h$.

Validation data provide fixed prediction-error scales $\sigma$.
The relations $b_{h,m}=\kappa_\alpha\sigma_{h,m}$ and $b_{h,\psi}=\kappa_\alpha\sigma_{h,\psi}$ determine the widths.
Each calibration window uses $100\,\mathrm{Hz}$ samples over the $3\,\mathrm{s}$ horizon.
Its score is the maximum scale-normalized residual across all position directions, circular yaw errors, and sampled times.
The shared factor $\kappa_\alpha$ is the empirical $(1-\alpha)$ quantile of these scores.
Each calibration voyage receives equal total weight.
Linear position interpolation and shortest-arc yaw interpolation connect the current error to the predicted centers.
The widths use linear interpolation between prediction nodes.
Terminal centers and widths remain constant beyond $3\,\mathrm{s}$.

\subsection{Error-Driven Occupancy Deformation}
\label{subsec:residual_deformation}

Directly incorporating position and yaw tracking errors into clearance calculation introduces nonlinear coupling, which complicates the derivation of analytical gradients. To address this difficulty, we use a convex hull to capture the combined effects of both tracking errors on vessel occupancy. This representation simplifies the gradient derivation by converting the coupled clearance calculation into a geometric projection of the convex hull along obstacle directions.

We construct this polygonal enclosure by combining the yaw-swept hull with the position-error set, as shown in Fig.~\ref{fig:system_architecture}(b).
The set $\mathcal H=\operatorname{conv}\{\mathbf b_1,\ldots,\mathbf b_4\}$ represents the rectangular hull with the reference origin in its interior, where $\mathbf b_a$ denotes a hull vertex in the body frame.
The sets $\mathcal E_{h,\mathrm p}$ and $\mathcal E_{h,\psi}$ specify the predicted translation and yaw ranges.

Continuous yaw sweeps can produce curved boundaries.
Fixed directional bounds provide a polygonal enclosure:
\begin{equation}
\begin{aligned}
    s_{h,\ell}&=\max_{a=1,\ldots,4}\max_{\varepsilon\in\mathcal E_{h,\psi}}
      \mathbf w_\ell^T\mathbf R(\varepsilon)\mathbf b_a,\\
    \widehat{\mathcal H}_h
    &=\{\mathbf x:\mathbf w_\ell^T\mathbf x\leq s_{h,\ell},
      \ \ell=1,\ldots,N_{\mathrm{dir}}\}.
\end{aligned}
\label{eq:yaw_sweep_polygon}
\end{equation}
Here, $\mathbf R$ denotes planar rotation, and $\mathbf w_\ell$ denotes a fixed body-frame unit direction.
The directions include the four hull normals for boundedness.
Each support value $s_{h,\ell}$ is the largest corner projection over the full yaw range.
Its computation checks interval endpoints and interior alignment angles.
These candidates capture projection maxima between endpoint orientations.

The nominal heading rotates the swept polygon into world axes.
The Minkowski sum then adds the translation range:
\begin{equation}
    \mathcal K_h
    =\mathbf R(\psi_h)\widehat{\mathcal H}_h\oplus\mathcal E_{h,\mathrm p},
    \label{eq:rod_profile}
\end{equation}
where $\oplus$ denotes the Minkowski sum and $\psi_h$ denotes the trajectory-tangent heading~\cite{zheng_scan-planner_2026}.
The set $\mathcal K_h$ describes occupancy relative to the reference position $\mathbf p_h=\mathbf p(h\,\mathrm{s})$.
The world occupancy $\mathbf p_h+\mathcal K_h$ encloses all hull poses allowed by $\mathcal E_h$.
Convexification and finite directional bounds can enlarge this enclosure.
The product error set also omits position--yaw correlation.

Clearance rays originate at the predicted position $\mathbf p_h+\hat{\mathbf r}_h$.
The centered polygon $\overline{\mathcal K}_h=\mathcal K_h-\hat{\mathbf r}_h=\mathbf R(\psi_h)\widehat{\mathcal H}_h\oplus\mathcal U_h$ contains the origin.
Below, we omit the prediction index $h$ for local geometry.
For a compact convex set $\mathcal S$ containing the origin, $\rho_{\mathcal S}(\mathbf u)=\max\{\lambda\geq0:\lambda\mathbf u\in\mathcal S\}$ gives its extent along a unit ray $\mathbf u$.
Ray--boundary intersection computes the extent of $\overline{\mathcal K}$ for clearance evaluation.
The case $\mathcal E_{\psi}=\{0\}$ reduces occupancy to $\mathcal K=\mathbf R(\psi)\mathcal H\oplus\mathcal E_{\mathrm p}$.
An additional condition $\mathcal E_{\mathrm p}=\{\mathbf0\}$ recovers the nominal hull $\mathcal K=\mathbf R(\psi)\mathcal H$.

\subsection{Trajectory Optimization of Occupancy Deformation}
\label{subsec:deformable_trajectory_optimization}
Directly considering multiple possible vessel footprints requires repeated clearance and gradient calculations during trajectory optimization. This process increases computational costs and complicates the derivation of analytical gradients. To address this difficulty, we exploit the convex hull of the predicted vessel footprints to formulate directional clearance constraints. The convex hull preserves the maximum occupancy extent in any direction, allowing us to derive analytical gradients directly from its geometric transformation.

To formulate these clearance constraints, we construct a deformed occupancy hull along the candidate trajectory. Interpolated prediction centers and widths determine $\mathcal K(t)$ through Eq.~\eqref{eq:rod_profile}.
Clearance rays start at the predicted position $\mathbf p(t)+\hat{\mathbf r}(t)$.
The centered polygon $\overline{\mathcal K}(t)=\mathcal K(t)-\hat{\mathbf r}(t)$ provides the corresponding radial extent.
The clearance along ray $j$ is
\begin{equation}
\begin{aligned}
    d_j(t)={}&(\mathbf o_j-\mathbf p(t)-\hat{\mathbf r}(t))^T\mathbf u_j\\
    &-[\rho_{\overline{\mathcal K}}(\mathbf u_j)-r_{\mathrm{map}}]_+.
\end{aligned}
    \label{eq:ray_clearance}
\end{equation}
The vector $\mathbf u_j$ is the unit ray direction.
The point $\mathbf o_j$ is its obstacle anchor on the inflated map.
The radius $r_{\mathrm{map}}$ accounts for the existing map inflation, and $[x]_+=\max(x,0)$.
Each connected angular obstacle sector contributes its most restrictive ray.

The occupancy construction satisfies the enclosure constraint in Eq.~\eqref{eq:rod_problem}.
Penalty terms relax its motion and clearance inequalities into an unconstrained objective:
\begin{equation}
\begin{aligned}
    \min_{\mathbf q,\boldsymbol\tau}\quad
    J_{\mathrm{pen}}={}&J_{\mathrm{jerk}}+w_TT_\Sigma\\
    &+\int_0^{T_\Sigma}\left[\Phi_{\mathrm{dyn}}(t)
      +\sum_j\Phi_{\mathrm{obs}}(d_j(t))\right]dt.
\end{aligned}
\label{eq:deformable_unconstrained_problem}
\end{equation}
Here, $J_{\mathrm{jerk}}$ is the jerk integral, and the virtual variables $\boldsymbol\tau$ determine the positive segment durations $\mathbf T(\boldsymbol\tau)$.
The term $\Phi_{\mathrm{dyn}}$ penalizes motion-constraint violations.
The term $\Phi_{\mathrm{obs}}$ penalizes directional clearance below $\delta_{\mathrm h}$.
Temporal quadrature evaluates the integral.

Each inner solve fixes obstacle rays, prediction nodes, and calibrated widths.
The trajectory heading and evaluation times still change the occupancy geometry.
The collision-cost derivative separates predicted-position changes from shape changes:
\begin{equation}
\begin{aligned}
    \frac{\partial\Phi_{\mathrm{obs}}(d_j)}{\partial z}
    =\Phi_{\mathrm{obs}}'(d_j)\bigg[
    &-\mathbf u_j^T\frac{\partial(\mathbf p+\hat{\mathbf r})}{\partial z}\\
    &-\chi_j\frac{\partial\rho_{\overline{\mathcal K}}(\mathbf u_j)}{\partial z}
    \bigg],
\end{aligned}
\label{eq:collision_penalty_gradient}
\end{equation}
where $z$ denotes a scalar trajectory coordinate or evaluation time.
The factor $\chi_j$ is one above the inflation threshold and zero below it.
At equality, $\chi_j\in[0,1]$ represents a generalized derivative.
\begin{figure*}[!t]
\centering
\includegraphics[width=0.90\textwidth]{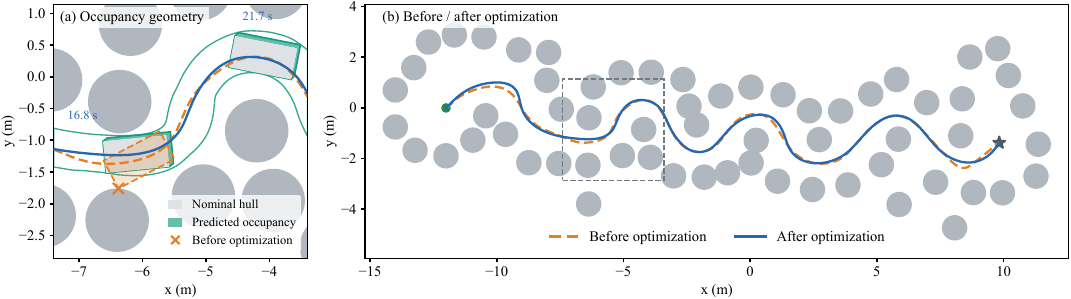}
\caption{Narrow-passage navigation before and after trajectory optimization. (a) Occupancy geometry and hull–obstacle contact before optimization. (b) Executed trajectories before and after optimization.}
\label{fig:goal_clearance}
\end{figure*}

An active polygon edge satisfies $\mathbf n_a^T\mathbf x=\beta_a$.
Its ray intersection gives $\rho_{\overline{\mathcal K}}(\mathbf u)=\beta_a/(\mathbf n_a^T\mathbf u)$ for $\mathbf n_a^T\mathbf u>0$.
Differentiation of this ratio propagates edge changes to the radial extent.
The edge derivatives follow the polygon vertices.
A retained vertex has the form $\mathbf x=\mathbf R(\psi)\mathbf W^{-1}\mathbf s+\mathbf c$, with $\mathbf c\in\operatorname{vert}(\mathcal U)$.
Its derivative is
\begin{equation}
    \frac{\partial\mathbf x}{\partial z}
    =\mathbf R(\psi)\mathbf A\mathbf y\frac{\partial\psi}{\partial z}
     +\mathbf R(\psi)\mathbf W^{-1}\frac{\partial\mathbf s}{\partial z}
     +\frac{\partial\mathbf c}{\partial z}.
    \label{eq:pose_envelope_vertex_gradient}
\end{equation}
The rows of $\mathbf W$ are two independent active support normals.
The vector $\mathbf s$ contains their support values.
The relation $\mathbf y=\mathbf W^{-1}\mathbf s$ gives the local vertex.
The matrix $\mathbf A=\left[\begin{smallmatrix}0&-1\\1&0\end{smallmatrix}\right]$ generates planar rotation.
The three terms describe nominal rotation, yaw-envelope changes, and position-uncertainty changes.
Active arc endpoints determine yaw-bound derivatives.
Interior maximizing angles and full-circle yaw sets contribute zero support derivative.

We solve each inner problem with LBFGS-Lite~\cite{wang_geometrically_2022} and refresh the predictions, bounds, and obstacle rays between solves.
We accept a trajectory only if its occupancy $\mathbf p(t)+\mathcal K(t)$ clears the raw obstacle map at all densely sampled times.

\section{Experimental Results}
\label{sec:experiments}

Simulation experiments and real-world experiments are conducted to validate the effectiveness and superiority of our methods.

\subsection{Experimental Setup}
\label{subsec:exp_setup}

We compare TEOD-Planner with EGO-Planner~\cite{zhou2021egoPlanner}, the USV adaptation of SCAN-Planner~\cite{zheng_scan-planner_2026}, and an ablation using separate position and yaw margins (Table~\ref{tab:planner_roles}).
\begin{table}[!htbp]
\centering
\caption{Planner Features and Comparison Roles}
\label{tab:planner_roles}
\footnotesize
\setlength{\tabcolsep}{3pt}
\begin{tabular*}{\columnwidth}{@{\extracolsep{\fill}}llll@{}}
\hline
Method & Hull model & Tracking error & Role \\
\hline
EGO & Fixed inflation & Not modeled & Basic baseline \\
SCAN & Yaw-aware & Not modeled & External comparison \\
Ours w/o joint def. & Yaw-aware & Separate margin & Coupling ablation \\
TEOD & Yaw-aware & Joint occupancy & Full method \\
\hline
\end{tabular*}
\end{table}

Each method runs on the same 30 scenes at each of three obstacle counts (50, 75, and 100), with identical start and goal positions and tracking controllers.
The ROS simulator runs on an Intel Core i7-12800HX CPU with a $1.0\,\mathrm{m}\times0.5\,\mathrm{m}$ USV, $0.1\,\mathrm{m}$ grid resolution, and a $7.5\,\mathrm{m}$ planning horizon.
All methods use speed and acceleration limits of $0.5\,\mathrm{m/s}$ and $1.0\,\mathrm{m/s^2}$.
A shared space- and time-varying flow field (up to $0.48\,\mathrm{m/s}$) drives position and yaw offsets bounded by $0.10\,\mathrm{m}$ and $8^\circ$.

Table~\ref{tab:dense_navigation} compares the planners on three metrics across the 30 scenes with 100 obstacles.
Success rate (SR) is the percentage of trials in which the vessel stays within $0.25\,\mathrm{m}$ of the goal for $1\,\mathrm{s}$ before the $180\,\mathrm{s}$ timeout without any hull--obstacle contact during the trial.
Collision rate (CR) is the percentage of trials with at least one hull--obstacle contact.
Mean planning-cycle time (Cycle) is computed by averaging the durations of all planning attempts, including rejected attempts, within each trial and then averaging the trial means.

\begin{table}[H]
\centering
\caption{Navigation Performance in the 100-Obstacle Scenes}
\label{tab:dense_navigation}
\footnotesize
\begin{tabular*}{\columnwidth}{@{\extracolsep{\fill}}lrrr@{}}
\hline
Method & SR (\%) $\uparrow$ & CR (\%) $\downarrow$ & Cycle (ms) $\downarrow$ \\
\hline
EGO-Planner & 46.7 & 53.3 & 0.71 \\
SCAN-Planner & 10.0 & 90.0 & 4.34 \\
Ours w/o joint def. & 80.0 & 3.3 & 16.22 \\
\hline
\textbf{TEOD-Planner (ours)} & \textbf{100.0} & \textbf{0.0} & \textbf{8.64} \\
\hline
\end{tabular*}
\end{table}

\begin{figure*}[!t]
\centering
\includegraphics[width=0.88\textwidth]{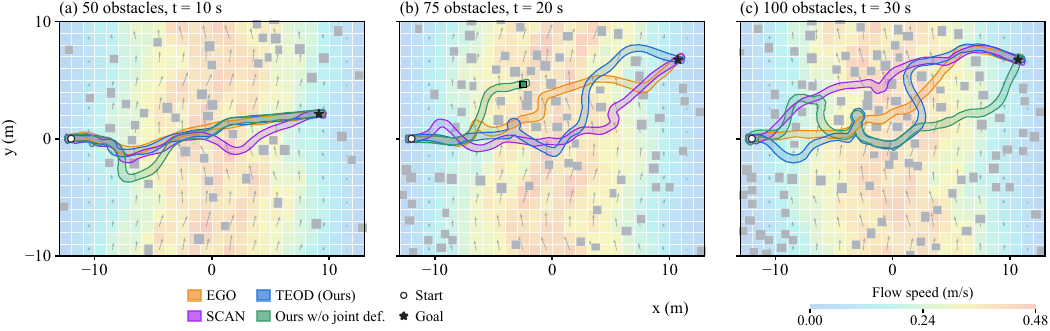}
\caption{Executed hull envelopes with (a) 50, (b) 75, and (c) 100 obstacles. Background grid colors indicate flow speed, and arrows indicate flow direction. The time-varying flow simulates river currents. Squares mark timeout endpoints.}
\label{fig:density_scenes}
\end{figure*}

\begin{figure}[!htbp]
\centering
\includegraphics[width=\columnwidth]{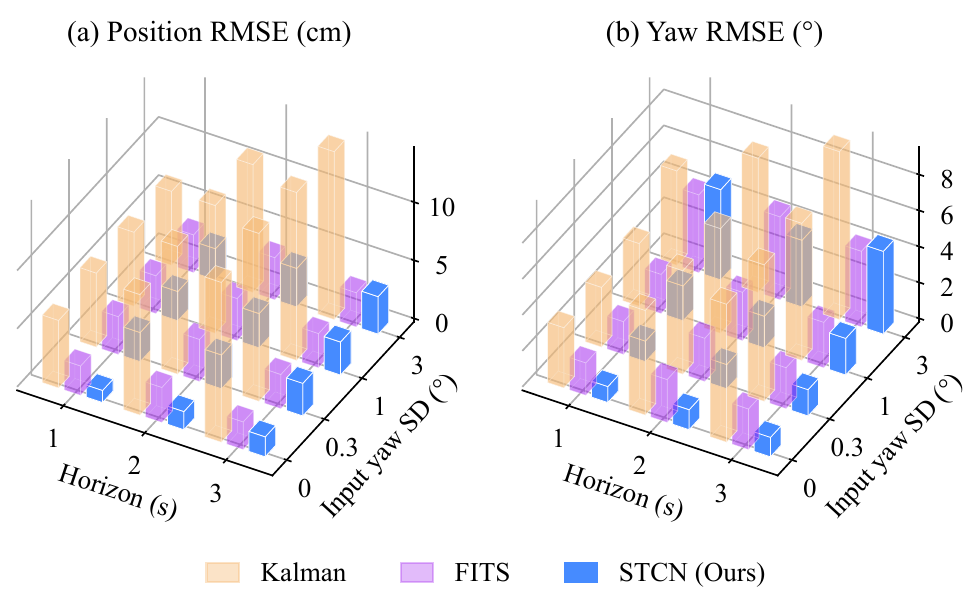}
\caption{Bar height gives mean prediction RMSE for (a) position and (b) yaw at each prediction horizon and input yaw noise level. The $0^\circ$ condition uses clean observations.}
\label{fig:stcn_prediction}
\end{figure}

\begin{figure}[!htbp]
\centering
\includegraphics[width=\columnwidth]{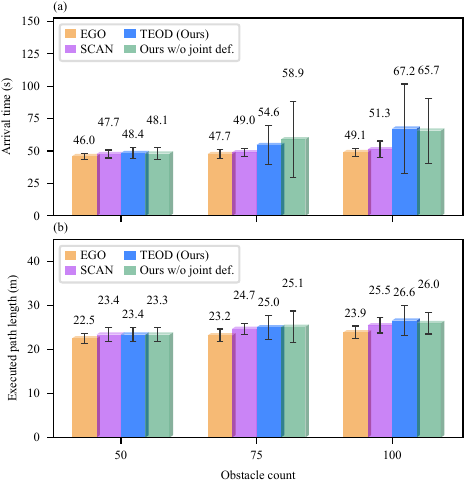}
\caption{Traversal costs for trials that reached the goal versus obstacle count. (a) Mean arrival time. (b) Mean executed path length. Error bars indicate sample standard deviations across scenes.}
\label{fig:density_planning_time}
\end{figure}

\begin{figure*}[!htbp]
\centering
\includegraphics[width=\textwidth]{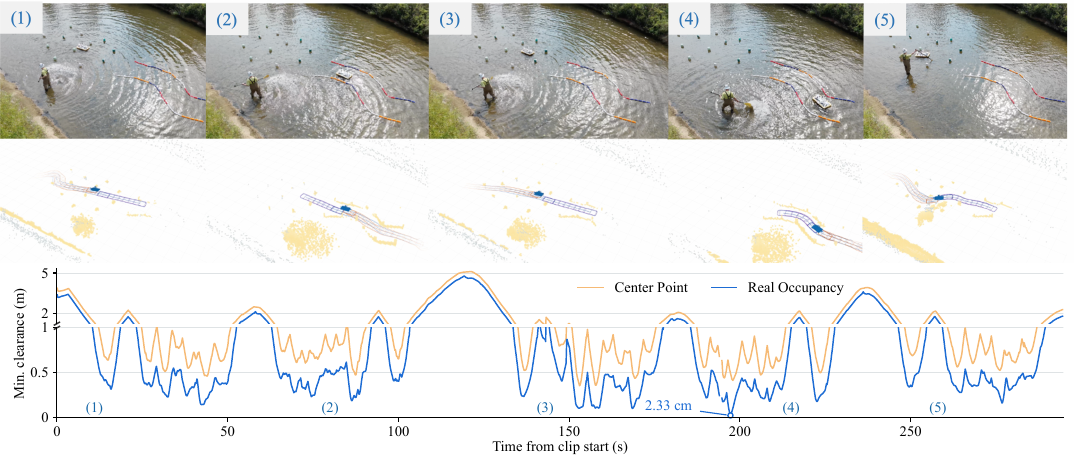}
\caption{Real-world confined navigation under continuous water-current disturbances. Top: experimental snapshots. Middle: onboard visualization. Bottom: minimum obstacle clearance of the vessel center and hull.}
\label{fig:real_exp_data}
\end{figure*}
\subsection{Tracking Error Prediction Accuracy}
\label{subsec:exp_prediction}

We evaluate STCN on 56 simulated voyages collected separately from the navigation benchmark (24 training, four validation, 20 calibration, and eight test voyages). We compare STCN with a constant-rate Kalman filter and a frequency-domain predictor adapted from the Frequency Interpolation Time Series Analysis Baseline (FITS)~\cite{xu_fits_2024}. Both baselines use the past $3\,\mathrm{s}$ of tracking errors. STCN also uses the upcoming reference trajectory. We fit FITS on clean and noise-augmented training voyages and select its frequency cutoff and the Kalman noise parameters on separate validation voyages across all noise levels. We evaluate the frozen predictors on 1,200 held-out windows using clean observations and three noise levels. In the noisy trials, we add Ornstein--Uhlenbeck noise with a correlation time of $0.3\,\mathrm{s}$, a standard deviation of $1\,\mathrm{cm}$ per position axis, and a yaw standard deviation of $0.3^\circ$, $1^\circ$, or $3^\circ$. All three predictors receive the same pose observations for each of three noise seeds.

Figure~\ref{fig:stcn_prediction} reports per-voyage RMSE against future observations, averaged across voyages and noise seeds. At the $3\,\mathrm{s}$ horizon, STCN had the lowest position and yaw RMSEs up to $1^\circ$ yaw noise. At $3^\circ$, FITS achieved the lowest errors ($2.93\,\mathrm{cm}$, $4.20^\circ$), followed by STCN ($3.22\,\mathrm{cm}$, $4.61^\circ$) and Kalman ($14.43\,\mathrm{cm}$, $9.34^\circ$). The envelope calibrated on clean data covered the observed errors throughout $95.0\%$ of clean windows but only $0.14\%$ of windows at $3^\circ$ yaw noise.

\subsection{Occupancy Deformation in Narrow Passages}
\label{subsec:exp_occupancy}

At the marked turn in Fig.~\ref{fig:goal_clearance}, the hull contacted an obstacle before optimization. As the vessel turns, a yaw error rotates its corners while a position error shifts the entire hull. The overlay in (a) illustrates how these effects change the clearance required on the obstacle side, using prescribed displacements at recorded positions. After optimization with this directional clearance, the executed trajectory passed the same region without contact (b).

\subsection{Navigation Comparison and Computational Cost}
\label{subsec:exp_density}

The $26\,\mathrm{m}\times20\,\mathrm{m}$ maps contain randomly placed obstacles, with each seed retaining the same goal across densities. In the three scenes shown in Fig.~\ref{fig:density_scenes}, TEOD reached every goal without hull contact. The separate-margin ablation also avoided contact in these examples but timed out in the 75-obstacle scene.

Across all 90 scenes, TEOD recorded zero hull contacts and reached 86 goals; all four timeouts occurred at the intermediate density. Counting each uninterrupted contact between the oriented hull and occupied cells as one event, we observed that contact events increased from 6 to 26 for EGO and from 16 to 72 for SCAN as obstacle count rose from 50 to 100. With 100 obstacles, TEOD achieved an SR of $100\%$, compared with $46.7\%$ for EGO and $10.0\%$ for SCAN (Table~\ref{tab:dense_navigation}). SCAN accounts for the hull at its nominal heading; the comparison shows the limitation of that clearance model under the tested tracking disturbances.

The ablation isolates the choice of clearance model. In the densest scenes, replacing separate position and yaw margins with joint occupancy deformation raised SR from $80\%$ to $100\%$ and reduced the fraction of trials with contact (CR) from $3.3\%$ to zero. Representing both errors through the hull geometry allowed the planner to retain traversable paths while accounting for clearance lost during tracking.

Joint deformation also reduced the computational cost of accounting for tracking errors. Its mean planning cycle increased from $4.94$ to $8.64\,\mathrm{ms}$ between 50 and 100 obstacles, compared with $5.21$ to $16.22\,\mathrm{ms}$ for separate margins, including inflation processing. At 100 obstacles, TEOD therefore used $46.7\%$ less planning time than the ablation (Table~\ref{tab:dense_navigation}). EGO and SCAN had shorter cycles, but both had a lower SR than TEOD. Among trials that reached the goal, TEOD and the ablation had similar mean path lengths ($26.6$ and $26.0\,\mathrm{m}$, respectively), as shown in Fig.~\ref{fig:density_planning_time}(b).

\begin{figure}[!htbp]
\centering
\includegraphics[width=0.95\columnwidth]{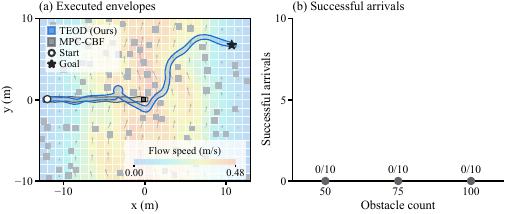}
\caption{MPC-CBF evaluation: (a) executed hull envelopes; (b) successful arrivals versus obstacle count.}
\label{fig:mpc_cbf_qualitative}
\end{figure}

In the tested passages, the STCN-informed MPC-CBF planner was too conservative to find a feasible trajectory through the obstacles within its planning time budget. The vessel therefore braked repeatedly and failed to clear the passage, as shown for the 75-obstacle scene in Fig.~\ref{fig:mpc_cbf_qualitative}(a). All 30 trials ended in timeout before the vessel reached the goal (Fig.~\ref{fig:mpc_cbf_qualitative}(b)).

\subsection{Real-World Confined Navigation}
\label{subsec:exp_real}

The test course combined an S-shaped passage with a cluster of fixed obstacles. Several gaps along the course were approximately \(1.5\) times the vessel width. With an operator paddling the water throughout one continuous experiment, the USV completed five consecutive traversals without contact or manual takeover (Fig.~\ref{fig:real_exp_data}). At its closest approach, the hull boundary at the measured vessel pose was $2.33\,\mathrm{cm}$ from a mapped obstacle. The larger center-point clearance would have overstated the space available at the hull.

\begin{figure}[!htbp]
\centering
\includegraphics[width=\columnwidth]{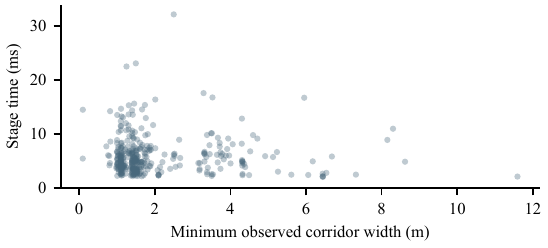}
\caption{Onboard trajectory optimization took approximately $5\,\mathrm{ms}$ on average.}
\label{fig:onboard_timing}
\end{figure}

On the Jetson Orin NX, trajectory optimization averaged $5.41\,\mathrm{ms}$ per published trajectory during the real-world experiment (Fig.~\ref{fig:onboard_timing}).

\section{Conclusion}
This paper presented TEOD-Planner for tracking-error-aware whole-body trajectory planning in narrow waterways.
Specifically, TEOD connects trajectory-conditioned tracking errors to hull occupancy and directional clearance.
First, the model encloses translated and rotated hulls within a polygon.
Then, exact radial extents and piecewise analytical gradients guide trajectory optimization.
Across 90 simulated trials, TEOD-Planner recorded zero hull--obstacle contacts during execution.
Mean planning-cycle times in simulation were below $5\,\mathrm{ms}$.
The physical USV demonstration showed passage through successive confined turns.
Future work will focus on predicting bounds on future position and yaw errors more accurately and improving trajectory tracking under stronger disturbances.

\ifCLASSOPTIONcaptionsoff
  \newpage
\fi

\bibliographystyle{Transactions-Bibliography/IEEEtran}
\bibliography{references}

\begin{thebibliography}{10}
\providecommand{\url}[1]{#1}
\csname url@samestyle\endcsname
\providecommand{\newblock}{\relax}
\providecommand{\bibinfo}[2]{#2}
\providecommand{\BIBentrySTDinterwordspacing}{\spaceskip=0pt\relax}
\providecommand{\BIBentryALTinterwordstretchfactor}{4}
\providecommand{\BIBentryALTinterwordspacing}{\spaceskip=\fontdimen2\font plus
\BIBentryALTinterwordstretchfactor\fontdimen3\font minus
  \fontdimen4\font\relax}
\providecommand{\BIBforeignlanguage}[2]{{%
\expandafter\ifx\csname l@#1\endcsname\relax
\typeout{** WARNING: IEEEtran.bst: No hyphenation pattern has been}%
\typeout{** loaded for the language `#1'. Using the pattern for}%
\typeout{** the default language instead.}%
\else
\language=\csname l@#1\endcsname
\fi
#2}}
\providecommand{\BIBdecl}{\relax}
\BIBdecl

\bibitem{barrera2021trends}
C.~Barrera, I.~Padr{\'o}n~Armas, F.~Luis, O.~Llinas, and G.~N.
  Marichal~Plasencia, ``Trends and challenges in unmanned surface vehicles
  ({USV}): From survey to shipping,'' \emph{TransNav}, vol.~15, no.~1, pp.
  135--142, Mar. 2021.

\bibitem{zhao2024optimal}
L.~Zhao, Y.~Bai, and J.~K. Paik, ``Optimal coverage path planning for
  {USV}-assisted coastal bathymetric survey: Models, solutions, and lake
  trials,'' \emph{Ocean Eng.}, vol. 296, Mar. 2024, {Art.} no. 116921.

\bibitem{vagale2021asvReview}
A.~Vagale, R.~Oucheikh, R.~T. Bye, O.~L. Osen, and T.~I. Fossen, ``Path
  planning and collision avoidance for autonomous surface vehicles {I}: A
  review,'' \emph{J. Mar. Sci. Technol.}, vol.~26, no.~4, pp. 1292--1306, Dec.
  2021.

\bibitem{zheng_scan-planner_2026}
H.~Zheng, Z.~Chen, Y.~Fu, M.~Yang, and T.~Qin, ``{SCAN-Planner}: Spatial
  collision-aware local planning for route-guided long-range quadruped
  navigation,'' 2026, {arXiv}:2606.19555.

\bibitem{lekkas2016ilos}
W.~Caharija \emph{et~al.}, ``Integral line-of-sight guidance and control of
  underactuated marine vehicles: Theory, simulations, and experiments,''
  \emph{{IEEE} Trans. Control Syst. Technol.}, vol.~24, no.~5, pp. 1623--1642,
  Sep. 2016.

\bibitem{gamboa-gonzalez_online_2025}
A.~Gamboa-Gonzalez, C.~Li, M.~Wehner, and W.~Wang, ``Online residual model
  learning for model predictive control of autonomous surface vehicles in
  real-world environments,'' in \emph{Proc. {IEEE/RSJ} Int. Conf. Intell.
  Robots Syst. ({IROS})}, Hangzhou, China, Oct. 2025, pp. 12\,712--12\,719.

\bibitem{park_robust_2023}
K.~Park, C.~Lee, and J.~Kim, ``Robust path tracking and obstacle avoidance of
  autonomous ship using stochastic model predictive control,'' in \emph{Proc.
  20th Int. Conf. Ubiquitous Robots ({UR})}, Honolulu, HI, USA, Jun. 2023, pp.
  179--182.

\bibitem{geng_robo-centric_2023}
S.~Geng, Q.~Wang, L.~Xie, C.~Xu, Y.~Cao, and F.~Gao, ``Robo-centric {ESDF}: A
  fast and accurate whole-body collision evaluation tool for any-shape robotic
  planning,'' in \emph{Proc. {IEEE/RSJ} Int. Conf. Intell. Robots Syst.
  ({IROS})}, Detroit, MI, USA, Oct. 2023, pp. 290--297.

\bibitem{son_3d_2026}
Y.~Son \emph{et~al.}, ``{3D RoA-Planner}: Path planner for quadruped robots in
  confined spaces using {3D} rotatable areas,'' \emph{{IEEE} Robot. Autom.
  Lett.}, vol.~11, no.~6, pp. 7436--7443, Jun. 2026.

\bibitem{van_der_saag_active_2025}
J.~van~der Saag, E.~Trevisan, W.~Falkena, and J.~Alonso-Mora, ``Active
  disturbance rejection control for trajectory tracking of a seagoing {USV}:
  Design, simulation, and field experiments,'' in \emph{Proc. {IEEE/RSJ} Int.
  Conf. Intell. Robots Syst. ({IROS})}, Hangzhou, China, Oct. 2025, pp.
  12\,720--12\,727.

\bibitem{duan_multimode_2025}
K.~Duan, S.~Dong, Z.~Fan, S.~Zhang, Y.~Shu, and M.~Liu, ``Multimode trajectory
  tracking control of unmanned surface vehicles based on {LSTM} assisted model
  predictive control,'' \emph{Ocean Eng.}, vol. 328, Jun. 2025, {Art.} no.
  121015.

\bibitem{walker_disturbance_2023}
K.~L. Walker and F.~Giorgio-Serchi, ``Disturbance preview for non-linear model
  predictive trajectory tracking of underwater vehicles in wave dominated
  environments,'' in \emph{Proc. {IEEE/RSJ} Int. Conf. Intell. Robots Syst.
  ({IROS})}, Detroit, MI, USA, Oct. 2023, pp. 6169--6176.

\bibitem{akbari2025currentGPMP}
B.~Akbari, Y.-J. Pan, S.~Liu, and T.~Wang, ``Optimized trajectory planning for
  {USVs} under ocean currents,'' \emph{{IEEE} J. Ocean. Eng.}, vol.~50, no.~4,
  pp. 2661--2671, Oct. 2025.

\bibitem{tagliabue2023robust}
A.~Tagliabue \emph{et~al.}, ``Robust, high-rate trajectory tracking on
  insect-scale soft-actuated aerial robots with deep-learned tube {MPC},'' in
  \emph{Proc. {IEEE} Int. Conf. Robot. Autom. ({ICRA})}, London, UK, May 2023,
  pp. 3383--3389.

\bibitem{wang_parametric_2025}
S.~Wang, M.~Zhu, Y.~Wen, and K.~Tian, ``Parametric dynamic modeling-based
  robust nonlinear model predictive control design for tracking unmanned
  surface vehicle trajectory,'' \emph{Ocean Eng.}, vol. 328, Jun. 2025, {Art.}
  no. 121074.

\bibitem{wang2024mpccbf}
W.~Wang, W.~Xiao, A.~Gonzalez-Garcia, J.~Swevers, C.~Ratti, and D.~Rus,
  ``Robust model predictive control with control barrier functions for
  autonomous surface vessels,'' in \emph{Proc. {IEEE} Int. Conf. Robot. Autom.
  ({ICRA})}, Yokohama, Japan, May 2024, pp. 6089--6095.

\bibitem{wang_geometrically_2022}
Z.~Wang, X.~Zhou, C.~Xu, and F.~Gao, ``Geometrically constrained trajectory
  optimization for multicopters,'' \emph{{IEEE} Trans. Robot.}, vol.~38, no.~5,
  pp. 3259--3278, Oct. 2022.

\bibitem{zhou2021egoPlanner}
X.~Zhou, Z.~Wang, H.~Ye, C.~Xu, and F.~Gao, ``{EGO-Planner}: An {ESDF}-free
  gradient-based local planner for quadrotors,'' \emph{{IEEE} Robot. Autom.
  Lett.}, vol.~6, no.~2, pp. 478--485, Apr. 2021.

\bibitem{zhou_swarm_2022}
X.~Zhou \emph{et~al.}, ``Swarm of micro flying robots in the wild,'' \emph{Sci.
  Robot.}, vol.~7, no.~66, May 2022, {Art.} no. eabm5954.

\bibitem{zhou_likelihood_2025}
Y.~Zhou and L.~Kong, ``Smooth likelihood-based collision avoidance for polygon
  shaped and differential drive vehicles,'' \emph{Int. J. Adv. Robot. Syst.},
  vol.~22, no.~4, pp. 1--14, Jul./Aug. 2025, {Art.} no. 17298806251352007.

\bibitem{jaitly_analytic_2025}
A.~Jaitly, D.~K. Jha, K.~Ota, and Y.~Shirai, ``Analytic conditions for
  differentiable collision detection in trajectory optimization,'' in
  \emph{Proc. IEEE/RSJ Int. Conf. Intell. Robots Syst. (IROS)}, Hangzhou,
  China, Oct. 2025, pp. 6659--6666.

\bibitem{gonzalez-garcia_safe_2025}
A.~Gonzalez-Garcia \emph{et~al.}, ``Safe motion planning and control using
  predictive and adaptive barrier methods for autonomous surface vessels,'' in
  \emph{Proc. {IEEE/RSJ} Int. Conf. Intell. Robots Syst. ({IROS})}, Hangzhou,
  China, Oct. 2025, pp. 5291--5298.

\bibitem{wang2018design}
W.~Wang \emph{et~al.}, ``Design, modeling, and nonlinear model predictive
  tracking control of a novel autonomous surface vehicle,'' in \emph{Proc.
  {IEEE} Int. Conf. Robot. Autom. ({ICRA})}, Brisbane, Australia, May 2018, pp.
  6189--6196.

\bibitem{xu_fits_2024}
Z.~Xu, A.~Zeng, and Q.~Xu, ``{FITS}: Modeling time series with 10k
  parameters,'' in \emph{Proc. Int. Conf. Learn. Represent. (ICLR)}, Vienna,
  Austria, May 2024, pp. 26\,295--26\,318.

\end{thebibliography}

\end{document}